\documentclass{article}

\usepackage[utf8]{inputenc} 
\usepackage[T1]{fontenc}    
\usepackage{hyperref}       
\usepackage{url}            
\usepackage{booktabs}       
\usepackage{amsfonts}       
\usepackage{nicefrac}       
\usepackage{microtype}      
\usepackage{graphicx}

\usepackage{natbib}
\usepackage[letterpaper, margin=1in]{geometry}

\title{Neural Architecture Design for \\GPU-Efficient Networks}

\usepackage{authblk}
\author[1]{Ming Lin\thanks{ming.l@alibaba-inc.com, https://minglin-home.github.io}}
\author[2]{Hesen Chen\thanks{hesen.chs@alibaba-inc.com}}
\author[2]{Xiuyu Sun\thanks{xiuyu.sxy@alibaba-inc.com}}
\author[1]{Qi Qian\thanks{qi.qian@alibaba-inc.com}}
\author[2]{Hao Li\thanks{lihao.lh@alibaba-inc.com}}
\author[1]{Rong Jin\thanks{jinrong.jr@alibaba-inc.com}}

\affil[1]{Alibaba Group, Bellevue, Washington, USA}
\affil[2]{Alibaba Group, Hangzhou, Zhejiang, China}

\begin{document}

\maketitle

\begin{abstract}
  Many mission-critical systems are based on GPU for inference. It requires not only high recognition accuracy but also low latency in responding time. Although many studies are devoted to optimizing the structure of deep models for efficient inference, most of them do not leverage the architecture of \textbf{modern GPU} for fast inference, leading to suboptimal performance. To address this issue, we propose a general principle for designing GPU-efficient networks based on extensive empirical studies. This design principle enables us to search for GPU-efficient network structures effectively by a simple and lightweight method as opposed to most Neural Architecture Search (NAS) methods that are complicated and computationally expensive. Based on the proposed framework, we design a family of GPU-Efficient Networks, or GENets in short. We did extensive evaluations on multiple GPU platforms and inference engines. While achieving $\geq 81.3\%$ top-1 accuracy on ImageNet, GENet is up to $6.4$ times faster than EfficienNet on GPU. It also outperforms most state-of-the-art models that are more efficient than EfficientNet in high precision regimes. Our source code and pre-trained models are available from \url{https://github.com/idstcv/GPU-Efficient-Networks}.

\end{abstract}

\section{Introduction}

Many mission-critical systems require high recognition accuracy within short response time. They are usually empowered by a high precision deep neural network and a modern GPU to shorten the inference latency. It is therefore important to leverage the architecture of \textbf{modern GPU} for fast inference.

Although there are many successful efficient networks with low inference latency \citep{cai_proxylessnas:_2019,tan_mnasnet:_2019,wu_fbnet:_2019,wan_fbnetv2_2020,cai_once-for-all_2020}, few take advantages of modern GPU, leading to suboptimal performance. The modern GPU has thousands of CUDA-cores and large built-in memory which make them insensitive to the network FLOPs (Floating Point Operations) and model size. Empirical benchmark \citep{radosavovic_designing_2020}  shows that the factors affecting the inference latency on modern GPUs are rather complicated. In this work, we aim to design an efficient \emph{high precision} network specially optimized for fast inference on modern GPU. Based on extensive empirical studies, we propose a general principle for designing GPU-efficient networks. This design principle  simplifies the the design space and therefore enables us to use a simple and lightweight Neural Architecture Search (NAS) method to search for GPU-efficient network effectively. Comparing to many brute-force NAS methods that are complicated and computationally expensive, the proposed design principle  improves the network efficiency on modern GPU within reasonable computational cost.

More specifically, we discover that on modern GPU, layers from different depths favor different convolutional operators. Low-level layers prefer full convolutions while high-level layers prefer depth-wise convolutions and bottleneck structure. Based on this observation, we propose a novel design principle which encourages a hybrid of different convolutional operators in one network. Then we use a lightweight but effective NAS algorithm to search for the GPU-Efficient Networks (GENets) under various latency budgets. The NAS algorithm generates three networks, GENet-light/normal/large, at different accuracy/speed trade-off points. We did extensive evaluations on multiple GPU platforms and inference engines. In general, \textbf{GENets are as accurate as EfficienNet-B2/B3 while being 6 times faster on modern GPU and are among the few state-of-the-art models that achieve $\geq 80.0\%$ top-1 accuracies on ImageNet}.

The NAS algorithm we used in this work is named \emph{Local-Linear-Regression NAS} (LLR-NAS). It learns a linear model to predict the accuracy of a given structure by perturbing an initial structure. Although we could use more powerful NAS algorithms in our design, we stick with LLR-NAS in this work. We wish to show that the superior performance of GENets is credited to a better understanding of the design space but rather credited to the machine brute-force NAS. We keep things as simple as possible. There are no SE block \citep{huSqueezeandExcitationNetworks2017}, transformers \citep{vaswaniAttentionAllYou2017} or even dropout \citep{srivastavaDropoutSimpleWay2014} to enhance our network. By embedding our design principle in our LLR-NAS, we significantly boost the inference speed  without bells and whistles. We summarize our main contributions as follows:
\begin{itemize}
  \item We propose a novel GPU-efficient design space which is highly optimized for fast GPU inference.
  \item Within our design space, we propose a lightweight but effective LLR-NAS to search for GPU-Efficient Networks. The resulting GENet-light/normal/large achieve high prediction accuracy on ImageNet while being several times faster than EfficientNet.
\end{itemize}

\section{Related Work}

Manually-designed networks usually use a single type of convolutions. ResNet-18 and ResNet-34 \citep{heDeepResidualLearning2016} use 3x3 full convolutions in all stages. ResNet-50 and deeper ones use bottleneck structure to reduce the computational cost and model size. The MobileNetV1/V2 \citep{howardMobileNetsEfficientConvolutional2017,sandlerMobileNetV2InvertedResiduals2018} replace full convolution with depth-wise convolution to further reduce the FLOPs. The emergence of NAS system greatly accelerates the design of efficient networks. A majority of NAS-generated networks use the same or similar design space as MobileNetV2, including EfficientNet \citep{tanEfficientNetRethinkingModel2019}, MobileNetV3 \citep{howardSearchingMobileNetV32019}, FBNet \citep{wu_fbnet:_2019,wan_fbnetv2_2020}, DNANet\citep{liBlockwiselySupervisedNeural2019}, OFANet \citep{cai_once-for-all_2020} and so on. The MixNet \citep{tanMixConvMixedDepthwise2019} proposed to hybridize depth-wise convolutions of different kernel size in one layer. More earlier works such as DART \citep{liuDARTSDifferentiableArchitecture2018} considered a hybrid of full, depth-wise and dilated convolution in one cell. This often results in irregular cell structures that are not GPU-friendly. The \citet{radosavovic_designing_2020} did large-scale experiments to train networks consisting of different bottleneck ratio $r$ and grouping parameter $g$. The resulting RegNet shares $r$ and $g > 1$ in all stages. In comparison, GENet uses $r\leq 1, g=1$ in low-level stages and $r\geq 1, g=\textrm{channels}$ in high-level stages.

\section{Network Design Space}
\label{sec:Network_Design_Space}

In this section, we describe how we came to our network design space. We first introduce three basic convolutional blocks widely used in popular networks. Then we benchmark the inference latency of each basic block and plot the singular value distribution of convolutional kernels. These profiling experiments suggest that we should use different basic blocks in different stages.

\subsection{Basic Convolutional Blocks and MasterNet Backbone}

\begin{figure}
  \centering
  \includegraphics{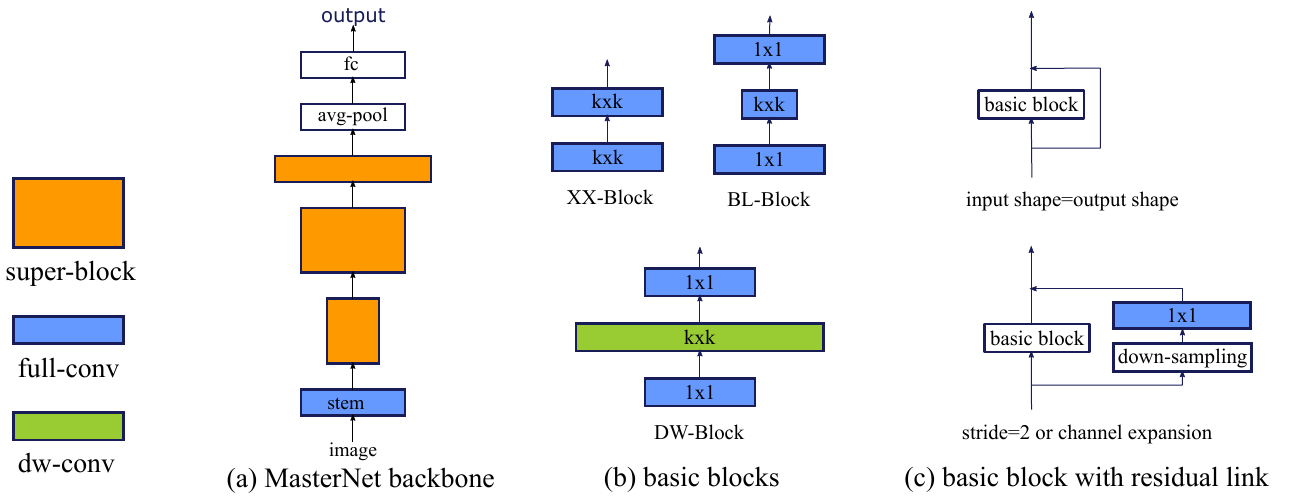}
  \caption{MasterNet backbone and basic blocks} \label{fig:basic_blocks}
\end{figure}

We focus on regularized structural space where the main body of the network consists of several super-blocks without skip-link, as shown in Figure \ref{fig:basic_blocks}(a). The network is called MasterNet. Each super-block is stacked by several basic blocks of the same type and width. Then a super-block is parameterized by its depth, width and stride. The first basic block in a super-block may be (but not necessarily) a down-sampling layer of stride $s=2$. The remaining basic blocks in the same super-block must be of stride $s=1$. The first layer of the MasterNet is a stem layer with stride $s=2$. After the main body, an average-pooling layer is followed by a fully-connected layer. Albeit rudimentary, this regularized structural space is good enough to design high precision low-latency networks on GPU.

We consider three basic blocks in Figure \ref{fig:basic_blocks}(b): XX-Block, BL-Block, DW-Block. 
The blue block represents full convolutional layer. The green block represents depth-wise convolutional layer. The BL-Block is a full convolutional layer with bottleneck structure of ratio $r \leq 1$. The DW-Block is a depth-wise convolutional layer with inverted bottleneck structure of ratio $r \geq 1$. All convolutional layers are followed by batch-normalization and RELU. For stride $s=2$ blocks, the down-sampling occurs at the first $k > 1$ layer. For clarity they are not shown in Figure \ref{fig:basic_blocks}. A basic block is always wrapped inside a residual structure. If the input and the output shape do not match, the identity link is replaced by a projection layer with down-sampling (if stride $s=2$), as shown in Figure \ref{fig:basic_blocks}(c).

Sometimes it is convenient to define the width of DW-Block to be the width of the depth-wise convolutional layer. Then the DW-Block consists of a depth-wise convolution followed by two $1\times 1$ bottleneck projection layers. The new bottleneck ratio $r_\mathrm{new}=1/r$. In this way, we uniformly parameterize BL-Block and DW-Block by their width $c$ and bottleneck ratio $r$.

\subsection{GPU Inference Latency}

We design controlled experiments to benchmark the inference speed of the three basic blocks on GPU. All timing experiments are repeated 30 times. The highest and the lowest $10\%$ numbers are removed and then the averaged inference latency is reported. We run benchmark experiments on NVIDIA V100 at half precision (FP16) and image resolution 224x224.

\paragraph{Batch Size}

\begin{figure}
  \hfil %
  \begin{minipage}{0.24\linewidth}
    \centering
    \includegraphics[width=\linewidth]{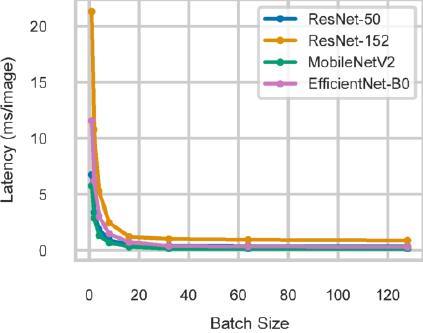} \\
    (a)
  \end{minipage} \hfil %
  \begin{minipage}{0.24\linewidth}
    \centering
    \includegraphics[width=\linewidth]{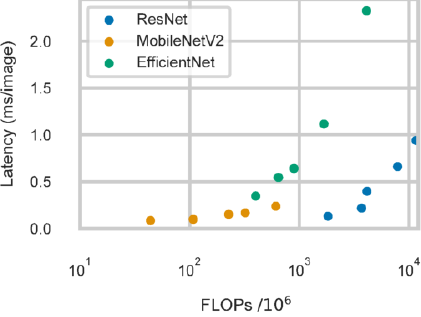} \\
    (b)
  \end{minipage} \hfil %
  \begin{minipage}{0.24\linewidth}
    \centering
    \includegraphics[width=\linewidth]{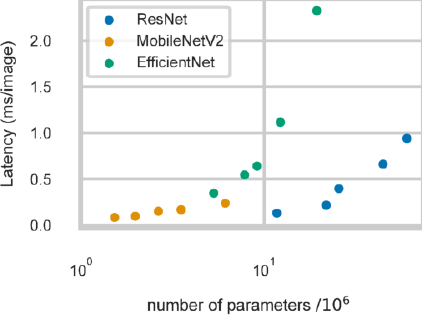} \\
    (c)
  \end{minipage}\hfil %
  \caption{Network latency, batch size 64, V100, FP16.} \label{fig:network_latency}
\end{figure}

In Figure \ref{fig:network_latency}(a), we vary the batch size from 1 to 128 and measure the inference latency of ResNet-50, ResNet152, MobileNetV2 and EfficientNet-B0. The y-axis is the latency of each network measured by milliseconds per image. That is, the inference latency of one mini-batch divided by the batch size. In the figure, the inference latency is not a constant value but a decaying function of batch size. This is because a large batch size allows GPU to do the computation parallelly thus more efficiently. Due to this reason, we must specify the batch size when comparing the inference latencies of different networks. In this work we focus on the setting of batch size $B=64$, a value adopted by many previous works \citep{cai_once-for-all_2020,radosavovic_designing_2020}. 

\paragraph{FLOPs and Model Size}

The FLOPs is a popular index to measure the network efficiency on edge device. We test whether the inference latency is still in proportion to the FLOPs on GPU. In Figure \ref{fig:network_latency}(b), we show the inference latency against FLOPs for ResNet, MobileNetV2 and EfficientNet at batch size $B=64$. We show the same scatter plot for model size in Figure \ref{fig:network_latency}(c). It is clear that the inference latency on GPU depends on neither FLOPs nor model size at all. This phenomenon was also observed in \citep{cai_once-for-all_2020,radosavovic_designing_2020}.

\paragraph{Basic Blocks}

\begin{figure}
  \begin{minipage}{0.24\linewidth}
    \centering
    \includegraphics[width=\linewidth]{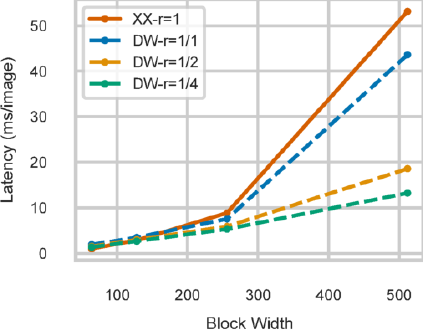} \\
    (a)
  \end{minipage} \hfil %
  \begin{minipage}{0.24\linewidth}
    \centering
    \includegraphics[width=\linewidth]{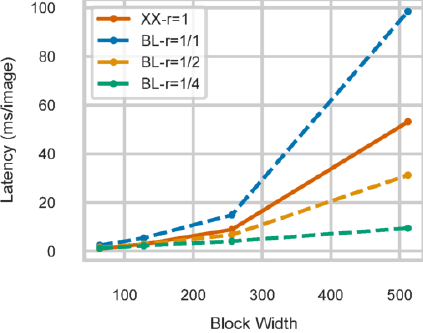} \\
    (b)
  \end{minipage} \hfil %
  \begin{minipage}{0.24\linewidth}
    \centering
    \includegraphics[width=\linewidth]{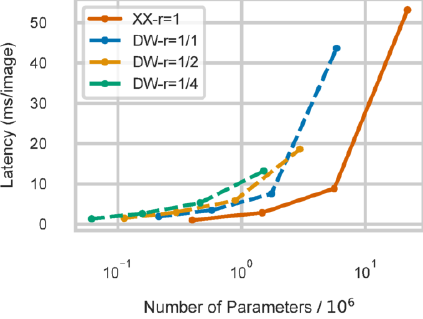} \\
    (c)
  \end{minipage} \hfil %
  \begin{minipage}{0.24\linewidth}
    \centering
    \includegraphics[width=\linewidth]{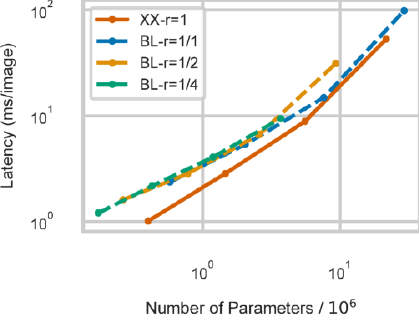} \\
    (d)
  \end{minipage}
  \caption{Basic block latency, resolution 224, batch size 64, V100, FP16.} \label{fig:basic_block_latency}
\end{figure}

Now we profile the three basic blocks. As mentioned before, we define the width of DW-Block to be the number of channels of the depth-wise convolutional layer and then its bottleneck ratio $r$ is re-defined to be $1/r$. We build three profiling networks only using one of the three basic blocks with the same width respectively. We set stride $s=1$, kernel size $k=3$ for all blocks. To benchmark XX-Block, the profiling network consists of 5 XX-Blocks. For BL-Block and DW-Block, we use 10 instead of 5 blocks to ensure the same visual region among networks. Then we sample $64$ images of resolution 224x224 and feed them into the profiling network. The inference latency is shown in Figure \ref{fig:basic_block_latency}.

In Figure \ref{fig:basic_block_latency}(a), we compare XX-Block against DW-Block under various $r$ and block widths. It is not surprising that with smaller $r$, the inference latency of DW-Block is smaller than XX-Block. We compare XX-Block against BL-Block under the same setting in Figure \ref{fig:basic_block_latency}(b). When $r=1$, BL-Block is slower than XX-Block since two BL-Blocks of $r=1$ is clearly larger than one XX-Block.

Based on Figure \ref{fig:basic_block_latency}(a)(b), it seems reasonable to only use BL-Block or DW-Block in network design, a strategy adopted by ResNet and EfficientNet. However, we argue that there is one more dimension to consider --- the network capacity. The network capacity affects its ability of fitting complex functions. One simple measurement of the network capacity is the number of free parameters of the network, also known as the model size. In Figure \ref{fig:basic_block_latency}(c) and (d)), we plot the network latency against the network size. We find that XX-Block is actually very efficient under the same model size.

Informally speaking, both BL-Block and DW-Block are kind of low-rank approximation of XX-Block. The above profiling results show that if we approximate XX-Block with BL/DW-Block, the approximation loss is in proportion to the intrinsic rank of the original XX-Block. If the XX-Block is nearly full rank, we have to set $r=1.0$ in BL/DW-Block which is actually inefficient. Therefore, \textbf{whether BL/DW-Block is more efficient than XX-Block on GPU depends on whether a particular layer prefers low-rank structure or not.}

\subsection{Intrinsic Rank of Convolutional Layers}

\begin{figure}
  \hfil %
  \begin{minipage}{0.24\linewidth}
    \centering
    \includegraphics[width=\linewidth]{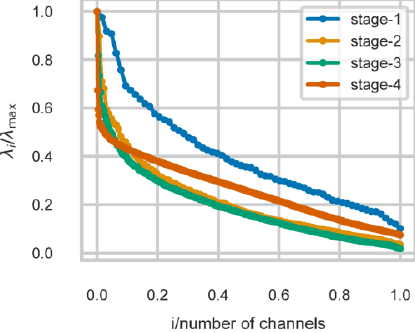} \\
    (a) ResNet-18
  \end{minipage} \hfil %
  \begin{minipage}{0.24\linewidth}
    \centering
    \includegraphics[width=\linewidth]{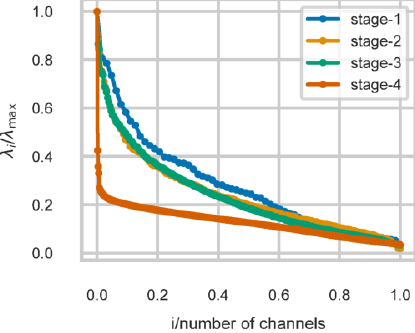} \\
    (b) ResNet-34
  \end{minipage} \hfil %
  \begin{minipage}{0.24\linewidth}
    \centering
    \includegraphics[width=\linewidth]{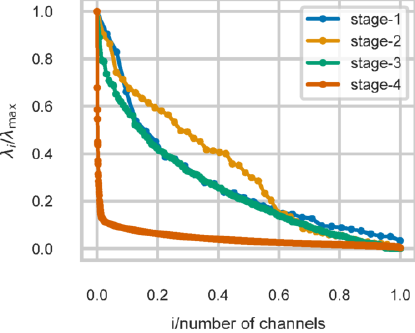} \\
    (c) ProfilingNet-132
  \end{minipage}\hfil %
  \caption{Singular value distribution of convolutional kernels.} \label{fig:svd_dist_resnets}
\end{figure}


Inspired by the profiling results in the previous subsection, we analyze the intrinsic rank of convolutional kernels in different stages. To this end, we inspect three networks consisting of only XX-Blocks. The first two are ResNet-18 and ResNet-34. They both use only XX-Blocks. ResNet-50 or deeper ResNets use bottleneck structure therefore are excluded. The third network is our manually designed network named ProfilingNet-132. It contains 132 convolutional layers of kernel size $k=3$ (see appendix for details). All three networks are trained on ImageNet up to 360 epochs. We plot the singular value distribution of convolutional kernels from different stages in Figure \ref{fig:svd_dist_resnets}. In each stage, we extract the convolutional kernel from the last layer. Excluding the first stem layer, there are 4 stages in total. Then we reshape the convolutional kernel to be a matrix of shape $(c_\mathrm{out}, c_\mathrm{in} \times k^2)$ where $c_\mathrm{out}, c_\mathrm{in}$ are the numbers of output channels and input channels respectively; $k$ is the kernel size which is equal to $3$ here. The singular values of the reshaped convolutional kernel matrix is computed. The largest singular value is normalized to $1.0$. For the $i$-th singular value $\lambda_i$, its x-coordinate is normalized to $i/c_\mathrm{out}$. From Figure \ref{fig:svd_dist_resnets}(a)-(c), it is clear that the singular values of high-level layers decay much faster than low-level ones. Therefore, the intrinsic rank of a layer decays along its depth. This means \textbf{we should use XX-Blocks in low-level stages and BL/DW-Blocks in high-level stages to maximize the GPU efficiency.}

\section{GPU-Efficient Networks}

\begin{table}
  \centering
  \caption{Manually designed networks (best 4 networks)}
  \label{tab:manumal_network}
  \begin{tabular}{llllll}
    \toprule
     model &     block type &       \# layers &                   \# channels &         stride &    Acc \\
    \midrule  
      Net1 &  C,X,X,B,D,D,C &  1,1,4,8,4,2,1 &    32,64,96,512,320,320,1280 &  2,2,2,2,2,1,1 &  77.6\% \\
      Net2 &  C,X,X,D,D,D,C &  1,1,4,8,4,2,1 &    32,48,64,160,320,320,1280 &  2,2,2,2,2,1,1 &  77.6\% \\
      Net3 &    C,X,X,D,D,C &    1,1,4,8,6,1 &        32,48,64,160,320,1280 &    2,2,2,2,2,1 &  77.5\% \\
      Net4 &    C,X,B,B,D,C &    1,1,4,8,6,1 &       32,64,256,512,320,1280 &    2,2,2,2,2,1 &  77.4\% \\  
    \bottomrule
    \end{tabular}
\end{table}

From Section \ref{sec:Network_Design_Space}, we learn that a GPU-efficient network should use XX-Block in the low-level layers and BL/DW-Block in high-level layers. In this section, we use this prior knowledge to design our GPU-Efficient Network. First, we manually design our MasterNet. Then we use MasterNet to initialize LLR-NAS to further optimize its structure.

\subsection{Design MasterNet}

To obtain a good initial structure, we designed 20 networks. All networks are designed to have latency around 0.34ms per image at batch size 64. This is also the latency of ResNet-50 and EfficientNet-B0. We train the 20 networks on ImageNet up to 120 epochs. For BL-Block we fix $r=1/4$. For DW-Block, we fix $r=6$. The results are summarized in Table \ref{tab:manumal_network}. We only report the best 5 networks. Full table is given in appendix.

In Table \ref{tab:manumal_network}, the 1st column is the IDs of the 20 networks. The 2nd column is the type of basic blocks in each super-block. 'C' denotes a single convolutional layer. 'X' denotes XX-Block. 'B' denotes BL-Block. 'D' denotes DW-Block. The 3rd column is the number of layers of each super-block. The 4th column is the width of each super-block. The 5th column is the stride. The last column reports the top-1 accuracy on ImageNet. The kernel size of the first and the last block are 3 and 1 respectively. The other blocks use kernel size $k=5$.

Table \ref{tab:manumal_network} validates our profiling results. All best networks use XX-Blocks in low-level layers and use BL/DW-Blocks in high-level layers, exactly as we expected. We choose Net1 as our MasterNet. Please note that the exact configuration of the MasterNet does not matter a lot since we will introduce NAS later. Net2 and Net3 are also good choices of MasterNet.

\subsection{Local Linear Regression NAS}

We use our automated NAS engine to optimize the structure of the MasterNet. Our NAS algorithm is named \emph{Local Linear Regression NAS} (LLR-NAS). Suppose that the MasterNet has $M$ super-blocks $S_1, S_2,\cdots,S_M$. Each super-block $S_i$ can be parameterized by $S_i(d_i,c_i,T_i)$ where $d_i$ is the depth, $c_i$ is the block width (number of output channels) and $T_i$ is the block-type. In addition to the three basic block types, we treat blocks of different kernel size $k$ and bottleneck ratio $r$ as different block types. The LLR-NAS performs the following steps:
\begin{description}
  \item[Distillation] For $i={1,2,\cdots,M}$, replace $S_i$ with a random block $\hat S_{i}^j(d_i^j,c_i^j,T_i^j)$.  Fine-tune $\hat S_i^j$ until convergent. Obtain the validation accuracy $A_i^j$ of the new network structure. Repeat this step until each $S_i$ of type $T$ has collected $N$ fine-tuned accuracies $\{ A_i^1, A_2^1, \cdots, A_i^N \}$.
  \item[Regression] Denote $A^*$ as the accuracy of the MasterNet. For each $i$ and each block type $T$, use least square regression to estimate the pseudo-gradient $\mathbf{g}=[g_1,g_2]$ such that $g_1 (d_i^j-d_i) + g_2 (c_i^j - c_i) \approx A_i^j - A^* $. With pseudo-gradient $\mathbf{g}$, we are able to predict the accuracy of any network structure given its structural parameters $\{(d_i,c_i,T_i)\}_{i=1}^M$.
  \item[Selection] Randomly generate many structures $\{(d_i,c_i,T_i)\}_{i=1}^M$ and predict their accuracies using the pseudo-gradient $\mathbf{g}$. Choose the best structure within the latency budget.
\end{description}

\subparagraph{Searching for Resolution} We use the same pseudo-gradient $\mathbf{g}$ at different resolutions. In the Selection step, we select the best structures under different resolutions independently. Then the selected structures are trained at their corresponding resolutions. Finally the best resolution is the one achieving the highest accuracy after training.

In LLR-NAS, we set the latency budgets to be 0.34/0.2/0.1 ms per image on V100 FP16 at batch size $B=64$. The corresponding NAS-structures are named GENet-large/normal/light.

\section{Experiments}
\begin{table}
  \centering
  \caption{GENet structures.}
  \label{tab:tab_GENet_structure}
  \begin{tabular}{cccccccccccccccc}
    \toprule 
     & \multicolumn{5}{c}{light} & \multicolumn{5}{c}{normal} & \multicolumn{5}{c}{large}\tabularnewline
    \midrule 
    type & d & c & s & k & r & d & c & s & k & r & d & c & s & k & r\tabularnewline
    \midrule
    \midrule 
    Conv & 1 & 13 & 2 & 3 & 1 & 1 & 32 & 2 & 3 & 1 & 1 & 32 & 2 & 3 & 1\tabularnewline
    \midrule 
    XX & 1 & 48 & 2 & 3 & 1 & 1 & 128 & 2 & 3 & 1 & 1 & 128 & 2 & 3 & 1\tabularnewline
    \midrule 
    XX & 3 & 48 & 2 & 3 & 1 & 2 & 192 & 2 & 3 & 1 & 2 & 192 & 2 & 3 & 1\tabularnewline
    \midrule 
    BL & 7 & 384 & 2 & 3 & 0.25 & 6 & 640 & 2 & 3 & 0.25 & 6 & 640 & 2 & 3 & 0.25\tabularnewline
    \midrule 
    DW & 2 & 560 & 2 & 3 & 3 & 4 & 640 & 2 & 3 & 3 & 5 & 640 & 2 & 3 & 3\tabularnewline
    \midrule 
    DW & 1 & 256 & 1 & 3 & 3 & 1 & 640 & 1 & 3 & 3 & 4 & 640 & 1 & 3 & 3\tabularnewline
    \midrule 
    Conv & 1 & 1920 & 1 & 1 & 1 & 1 & 2560 & 1 & 1 & 1 & 1 & 2560 & 1 & 1 & 1\tabularnewline
    \midrule
    \midrule 
    FLOPs & \multicolumn{5}{c}{552 M@192x192} & \multicolumn{5}{c}{2.2 G@192x192} & \multicolumn{5}{c}{4.6 G@256x256}\tabularnewline
    \midrule 
    Params & \multicolumn{5}{c}{8.17 M} & \multicolumn{5}{c}{21 M} & \multicolumn{5}{c}{31 M}\tabularnewline
    \bottomrule
    \end{tabular}
\end{table}

\begin{figure}
  \hfil %
  \begin{minipage}{0.32\linewidth}
    \centering
    \includegraphics[width=\linewidth]{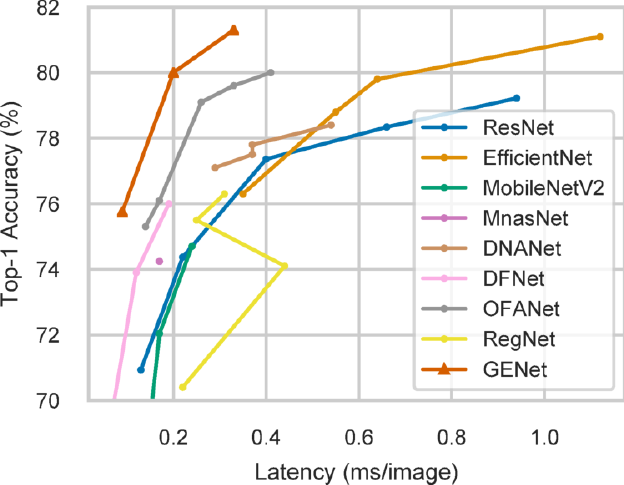}\\
    (a) V100 FP16
  \end{minipage} \hfil %
  \begin{minipage}{0.32\linewidth}
    \centering
    \includegraphics[width=\linewidth]{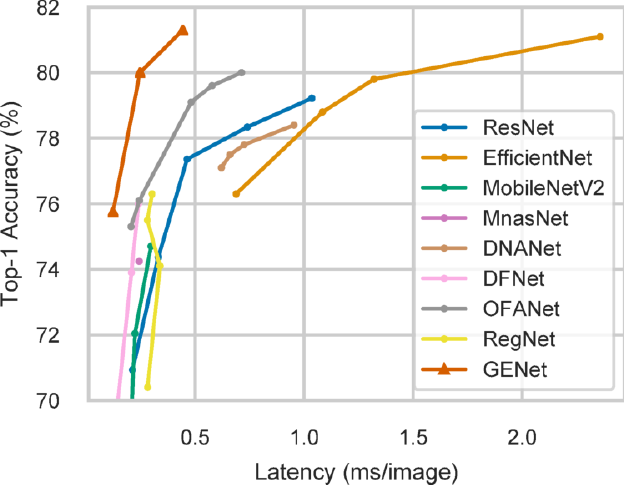}\\
    (b) T4 TensorRT FP16
  \end{minipage} \hfil %
  \begin{minipage}{0.32\linewidth}
    \centering
    \includegraphics[width=\linewidth]{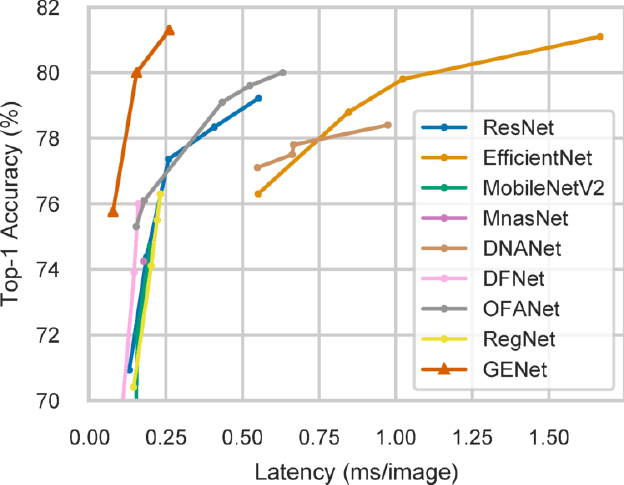}\\
    (c) T4 TensorRT INT8
  \end{minipage} \hfil %
  
  \caption{ImageNet top-1 accuracy v.s. network latency, batch size 64 on V100, batch size 32 on T4.} \label{fig:network_acc_vs_latency_and_fps}
\end{figure}

\begin{table}
  \centering
  \caption{GENet speed-up on V100 FP16 $B=64$ and T4 FP16/INT8 $B=32$.}
  \label{tab:tab_model_latency_speedup}
  \begin{tabular}{cccccccc}
    \toprule 
     &  & \multicolumn{2}{c}{V100 FP16} & \multicolumn{2}{c}{T4 FP16} & \multicolumn{2}{c}{T4 INT8}\tabularnewline
    \midrule 
    model & acc & latency & speed-up & latency & speed-up & latency & speed-up\tabularnewline
    \midrule
    \midrule 
    \textbf{GENet-light} & \textbf{75.7\%} & \textbf{0.1} & \textbf{1.0x} & \textbf{0.14} & \textbf{1.0x} & \textbf{0.1} & \textbf{1.0x}\tabularnewline
    \midrule 
    ResNet-34 & 74.4\% & 0.22 & 2.2x & 0.33 & 2.3x & 0.19 & 1.9x\tabularnewline
    \midrule 
    RegNetY-600MF & 75.5\% & 0.31 & 3.1x & 0.28 & 2.0x & 0.22 & 2.2x\tabularnewline
    \midrule 
    MobileNetV2-1.4 & 74.7\% & 0.24 & 2.4x & 0.30 & 2.1x & 0.20 & 2.0x\tabularnewline
    \midrule 
    DNANet-2a & 76.0\% & 0.19 & 1.9x & 0.24 & 1.7x & 0.16 & 1.6x\tabularnewline
    \midrule 
    OFANet-9ms & 75.3\% & 0.14 & 1.4x & 0.21 & 1.5x & 0.15 & 1.5x\tabularnewline
    \midrule
    \midrule 
    \textbf{GENet-normal} & \textbf{80.0\%} & \textbf{0.20} & \textbf{1.0x} & \textbf{0.25} & \textbf{1.0x} & \textbf{0.15} & \textbf{1.0x}\tabularnewline
    \midrule 
    ResNet-152 & 79.2\% & 0.94 & 4.7x & 1.04 & 4.2x & 0.55 & 3.7x\tabularnewline
    \midrule 
    EfficientNet-B2 & 79.8\% & 0.64 & 3.2x & 1.32 & 5.3x & 1.02 & 6.8x\tabularnewline
    \midrule 
    OFANet-595M & 80.0\% & 0.41 & 2.1x & 0.71 & 2.8x & 0.63 & 4.2x\tabularnewline
    \midrule
    \midrule 
    \textbf{GENet-large} & \textbf{81.3\%} & \textbf{0.34} & \textbf{1.0x} & \textbf{0.44} & \textbf{1.0x} & \textbf{0.26} & \textbf{1.0x}\tabularnewline
    \midrule 
    EfficientNet-B3 & 81.1\% & 1.12 & 3.3x & 2.36 & 5.4x & 1.67 & 6.4x\tabularnewline
    \bottomrule
    \end{tabular}
\end{table}

In this section, we compare GENet against popular networks, including ResNet \citep{heDeepResidualLearning2016}, EfficientNet \citep{tanEfficientNetRethinkingModel2019}, MobileNetV2 \citep{sandlerMobileNetV2InvertedResiduals2018}, MnasNet \citep{tan_mnasnet:_2019}, DNANet \citep{liBlockwiselySupervisedNeural2019}, DFNet \citep{liPartialOrderPruning2019}, OFANet \citep{cai_once-for-all_2020} and RegNet \citep{radosavovic_designing_2020}. We use PyTorch to implement our experiments. All timing experiments are averaged over 30 times with highest and lowest $10\%$ numbers removed.

We compare network accuracy on ImageNet \citep{deng_imagenet_2009}. We use the official ImageNet training/validation split. For LLR-NAS, we randomly sample 50000 images from the training set as NAS validation set which are not used in the LLR-NAS training. We use SGD with Nesterov momentum 0.9, initial learning rate 0.1 per 256 batch size, cosine learning rate decay, warm-up 5 epochs, weight-decay $4\times10^{-5}$. Parameters are initialized by MSRAPrelu \citep{10.1109/ICCV.2015.123}. In LLR-NAS, each perturbed super-block is fine-tuned up to 30 epochs with initial learning rate 0.01 with batch size 256. We tune kernel size in $\{3,5\}$; super-block width (number of output channels) in range $[2.0c, 0.5c]$ where $c$ is the width of the super-block; the depth of the super-block is tuned in $[d+2, d-2]$ where $L$ is the depth of the super-block. For BL-Block the bottleneck ratio $r$ is tuned in set $\{0.5,0.25\}$. For DW-Block, the $r$ is tuned in set $\{3,6,9\}$. For each super-block, we randomly sample 9 blocks within the above structural perturbation range. To select the best resolution, we tune input image resolution in $\{192,224,256\}$. At each resolution, we select a new network within the latency budget. Each network is trained 120 epochs. The overall computational cost of LLR-NAS is about 60 hours running on 24 V100 GPUs.  The final networks are trained up to 480 epochs with label-smoothing \citep{szegedyRethinkingInceptionArchitecture2016a}, mix-up \citep{zhangMixupEmpiricalRisk2017}, random-erase \citep{zhongRandomErasingData2017} and auto-augmentation \citep{cubukAutoAugmentLearningAugmentation2018}. We also use ResNet-152 as teacher network to get soft-labels. The soft-label loss and the true-label loss are weighted 1:1. Due to the space limitation, more details and results could be found in appendix.

Table \ref{tab:tab_GENet_structure} gives the structures of GENet-light/normal/large. The first layer is always the stem layer of stride $s=2$. Although the MasterNet uses kernel size $k=5$ in all layers, after NAS GENets choose $k=3$ consistently. The DW-Blocks choose $r=3$. This is very different to EfficientNet which chooses $r=6$ for all depth-wise blocks and $k=5$ for most blocks. Comparing 'light' to 'large', GENet-large increases the depth of high-level stages and reduces the depth of low-level stages. The last two rows give the FLOPs and model size of the GENets, as well as the preferred input image resolution.

To benchmark network inference speed on GPU, we considered three settings: 1) NVIDIA V100 using half precision (FP16) and PyTorch inference engine; 2) NVIDIA T4 FP16 and TensorRT inference engine; 3) NVIDIA T4 INT8 and TensorRT inference engine. The V100 + PyTorch is the most common setting in research papers. The T4 GPU and TensorRT engine are specially designed for inference service. We use the TensorRT built-in quantization tool to convert the networks to INT8 precision. We plot the accuracy of each network against its inference latency in Figure \ref{fig:network_acc_vs_latency_and_fps}. GENets outperform all baseline networks by a large margin. Particularly, GENet-large is $5.0\%$ more accurate than EfficienNet-B0 with the same speed on V100 and is the few state-of-the-art model achieving $81.3\%$ accuracy on ImageNet while being much faster than EfficienNet-B3. Although GENet is optimized for V100 GPU, it is even more efficient on T4 GPU. This shows the excellent transferable performance of GENet among different GPU platforms.

In Table \ref{tab:tab_model_latency_speedup}, we compare the inference latency of GENets against baseline networks around the similar top-1 accuracy. We could see that the superiority of GENets is more significant when higher accuracy is required. Most networks cannot achieve $\geq 81\%$ top-1 accuracy except EfficienNet-B3 and GENet-large. The T4 + TensorRT + INT8 configuration is widely used in industry. Under this setting, GENet-large is $6.4$ times faster than EfficienNet-B3 and GENet-normal is $6.8$ times faster than EfficienNet-B2. Comparing to the OFANet, GENet-normal is $4.2$ times faster at the $\mathrm{acc}=80.0\%$ level. Considering the fact that the second fastest OFANet is 1.6 times faster than EfficienNet-B2, our GENet is significantly faster than all baseline methods..

\section{Conclusion}

We propose a design space optimized for fast GPU inference. In this space, we use semi-automatic NAS to help us design GPU-Efficient Networks. GENets use full convolutions in low-level stages and depth-wise convolution and/or bottleneck structure in high-level stages. This design is inspired by the observation that convolutional kernels in the high-level stages are more likely to have low intrinsic rank and different types of convolutions have different kinds of efficiency on GPU. We wish this work sheds some light on the design principle of GPU-efficient networks and inspires more researchers to reconsider the efficiency of networks on modern GPU.

\bibliographystyle{plainnat}
\bibliography{refs}

\newpage
\appendix

\section{More Tables}
\label{sec:more_tables}

Table \ref{tab:llr-nas-diff-kernel-resolution} reports the top-1 accuracy on ImageNet for networks trained under different resolutions and kernel sizes. For 01ms latency budget, we fix kernel size $k=3$ and resolution as $192$ manually.

\begin{table}[h]
  \centering
  \caption{Top-1 accuracy under different kernel sizes and resolutions. }
  \label{tab:llr-nas-diff-kernel-resolution}
  \begin{tabular}{cccc}
    \toprule 
    latency (ms/image) & kerenl size & resoltuion & top-1 Accuracy\tabularnewline
    \midrule
    \midrule 
    0.34 & 5x5 & 192 & 78.9\%\tabularnewline
    \midrule 
     & 5x5 & 224 & 79.2\%\tabularnewline
    \midrule 
     & 5x5 & 288 & 78.9\%\tabularnewline
    \midrule 
     & 3x3 & 192 & 79.7\%\tabularnewline
    \midrule 
     & 3x3 & 224 & 79.7\%\tabularnewline
    \midrule 
     & 3x3 & 256 & 79.8\%\tabularnewline
    \midrule 
     & 3x3 & 288 & 79.1\%\tabularnewline
    \midrule 
    0.20 & 5x5 & 192 & 77.1\%\tabularnewline
    \midrule 
     & 5x5 & 224 & 77.5\%\tabularnewline
    \midrule 
     & 3x3 & 160 & 77.6\%\tabularnewline
    \midrule 
     & 3x3 & 192 & 78.2\%\tabularnewline
    \midrule 
     & 3x3 & 224 & 77.7\%\tabularnewline
    \midrule 
     & 3x3 & 224 & 77.7\%\tabularnewline
    \midrule 
     & 3x3 & 256 & 77.7\%\tabularnewline
    \bottomrule
\end{tabular}
\end{table}

Table \ref{tab:tab_profilingnet132_structure} gives the structure of ProfilingNet-132 in Section 3.3.
Table \ref{tab:manumal_network_full} gives the structures and accuracies of the 20 manually-designed networks in Section 4.1.

Table \ref{tab:Network-latency-on-V100-FP16-all-batchsize}, \ref{tab:Network-latency-on-T4-TensorRT-FP16-all-batchsize} and \ref{tab:Network-latency-on-T4-TensorRT-INT8-all-batchsize} report the accuracies and latencies of baseline networks and GENets on V100 and T4, with batch size from 1 to 64. 

On T4 we encounter GPU memory error when batch size is larger than 32 therefore only results up to 32 batch size on T4 are reported.

\begin{table}[hb]
  \centering
  \caption{ProfilingNet-132 structure.}
  \label{tab:tab_profilingnet132_structure}
  \begin{tabular}{llllll}
    \toprule
     type &  d &     c &  s &  k &    r \\
    \midrule
     Conv &  1 &    16 &  2 &  3 &  1.0 \\
       XX &  3 &    32 &  2 &  3 &  1.0 \\
       XX &  3 &    48 &  2 &  3 &  1.0 \\
       XX &  3 &    72 &  2 &  3 &  1.0 \\
       XX &  6 &   128 &  1 &  3 &  1.0 \\
       XX &  6 &   256 &  2 &  3 &  1.0 \\
       XX &  8 &   512 &  1 &  3 &  1.0 \\
       XX &  8 &  1024 &  1 &  3 &  1.0 \\
       XX &  4 &  2048 &  1 &  3 &  1.0 \\
     Conv &  1 &  4096 &  1 &  1 &  1.0 \\
    \bottomrule
    \end{tabular}    
\end{table}

\begin{table}[hb]
  \centering
  \caption{Manually designed networks (full table)}
  \label{tab:manumal_network_full}
  \begin{tabular}{llllll}
    \toprule
     Model &     Block Type &       \# Layers &                   \# Channels &         Stride &    Acc \\
    \midrule  
      Net1 &  C,X,X,B,D,D,C &  1,1,4,8,4,2,1 &    32,64,96,512,320,320,1280 &  2,2,2,2,2,1,1 &  77.6\% \\
      Net2 &  C,X,X,D,D,D,C &  1,1,4,8,4,2,1 &    32,48,64,160,320,320,1280 &  2,2,2,2,2,1,1 &  77.6\% \\
      Net3 &    C,X,X,D,D,C &    1,1,4,8,6,1 &        32,48,64,160,320,1280 &    2,2,2,2,2,1 &  77.5\% \\
      Net4 &    C,X,B,B,D,C &    1,1,4,8,6,1 &       32,64,256,512,320,1280 &    2,2,2,2,2,1 &  77.4\% \\
      Net5 &  C,X,B,D,D,D,C &  1,1,4,8,4,2,1 &   32,32,256,144,288,288,1280 &  2,2,2,2,2,1,1 &  77.4\% \\
      Net6 &    C,X,X,B,D,C &    1,1,4,8,6,1 &        32,64,96,512,320,1280 &    2,2,2,2,2,1 &  77.3\% \\
      Net7 &    C,X,B,D,D,C &    1,1,4,8,6,1 &       32,32,256,144,288,1280 &    2,2,2,2,2,1 &  77.3\% \\
      Net8 &  C,D,D,D,D,D,C &  1,1,4,8,4,2,1 &    32,24,64,128,256,256,1280 &  2,2,2,2,2,1,1 &  77.2\% \\
     Net9 &  C,X,X,D,D,D,C &  1,1,4,4,4,4,1 &    32,48,64,160,160,320,1280 &  2,2,2,2,1,2,1 &  77.1\% \\
     Net10 &    C,D,D,D,D,C &    1,1,4,8,6,1 &        32,24,64,128,256,1280 &    2,2,2,2,2,1 &  77.0\% \\
     Net11 &  C,X,X,D,D,D,C &  1,1,4,6,2,4,1 &    32,48,64,160,160,320,1280 &  2,2,2,2,1,2,1 &  76.9\% \\
     Net12 &  C,X,X,B,B,D,C &  1,1,4,6,2,4,1 &    32,64,96,512,512,320,1280 &  2,2,2,2,1,2,1 &  76.9\% \\
     Net13 &  C,X,X,B,B,D,C &  1,1,4,4,4,4,1 &    32,64,96,512,512,320,1280 &  2,2,2,2,1,2,1 &  76.8\% \\
     Net14 &  C,X,B,D,D,D,C &  1,1,4,6,2,4,1 &   32,32,256,144,144,288,1280 &  2,2,2,2,1,2,1 &  76.7\% \\
     Net15 &  C,X,X,B,B,B,C &  1,1,4,8,4,2,1 &  32,64,96,512,1024,1024,1280 &  2,2,2,2,2,1,1 &  76.6\% \\
     Net16 &  C,D,D,D,D,D,C &  1,1,4,6,2,4,1 &    32,24,64,128,128,256,1280 &  2,2,2,2,1,2,1 &  76.5\% \\
     Net17 &    C,X,B,B,B,C &    1,1,4,8,6,1 &      32,64,256,512,1024,1280 &    2,2,2,2,2,1 &  76.3\% \\
     Net18 &  C,X,X,B,B,B,C &  1,1,4,6,2,4,1 &   32,64,96,512,512,1024,1280 &  2,2,2,2,1,2,1 &  76.0\% \\
     Net19 &    C,X,X,B,B,C &    1,1,4,8,6,1 &       32,64,96,512,1024,1280 &    2,2,2,2,2,1 &  76.0\% \\
     Net20 &  C,X,X,B,B,B,C &  1,1,4,4,4,4,1 &   32,64,96,512,512,1024,1280 &  2,2,2,2,1,2,1 &  75.8\% \\
    \bottomrule
    \end{tabular}
    
\end{table}

\begin{table}[hb]
  \centering
  \caption{Network latency (ms/image) on V100, FP16, batch size 1 to 64.}
  \label{tab:Network-latency-on-V100-FP16-all-batchsize}
  \begin{tabular}{lllllllll}
    \toprule
                model &    acc &      1 &      2 &     4 &     8 &    16 &    32 &    64 \\
    \midrule
        RegNetY-200MF &  70.4\% &  13.19 &   7.02 &  3.33 &  1.74 &  0.84 &  0.51 &  0.22 \\
        RegNetY-400MF &  74.1\% &  16.42 &   8.22 &  4.88 &  2.08 &  1.13 &  0.53 &  0.44 \\
        RegNetY-600MF &  75.5\% &  13.61 &   7.74 &  3.39 &  1.71 &  0.85 &  0.45 &  0.25 \\
        RegNetY-800MF &  76.3\% &  12.80 &   6.69 &  3.69 &  1.96 &  0.82 &  0.46 &  0.31 \\
            ResNet-18 &  70.9\% &   2.87 &   1.39 &  0.71 &  0.38 &  0.19 &  0.15 &  0.13 \\
            ResNet-34 &  74.4\% &   5.13 &   2.42 &  1.23 &  0.62 &  0.32 &  0.24 &  0.22 \\
            ResNet-50 &  77.4\% &   6.77 &   3.42 &  1.86 &  0.93 &  0.46 &  0.43 &  0.40 \\
           ResNet-101 &  78.3\% &  13.24 &   7.41 &  3.62 &  1.67 &  0.83 &  0.71 &  0.66 \\
           ResNet-152 &  79.2\% &  21.33 &  10.81 &  5.30 &  2.48 &  1.23 &  1.01 &  0.94 \\
      EfficientNet-B0 &  76.3\% &  11.55 &   6.24 &  3.06 &  1.49 &  0.73 &  0.40 &  0.35 \\
      EfficientNet-B1 &  78.8\% &  16.55 &   8.30 &  4.09 &  2.07 &  1.03 &  0.59 &  0.55 \\
      EfficientNet-B2 &  79.8\% &  16.35 &   8.07 &  4.07 &  2.04 &  1.04 &  0.69 &  0.64 \\
      EfficientNet-B3 &  81.1\% &  18.64 &  10.04 &  5.06 &  2.39 &  1.28 &  1.18 &  1.12 \\
     MobileNetV2-0.25 &  51.8\% &   5.60 &   2.53 &  1.28 &  0.70 &  0.35 &  0.17 &  0.08 \\
      MobileNetV2-0.5 &  64.4\% &   5.14 &   2.85 &  1.29 &  0.65 &  0.33 &  0.17 &  0.10 \\
     MobileNetV2-0.75 &  69.4\% &   5.85 &   2.90 &  1.33 &  0.67 &  0.37 &  0.19 &  0.15 \\
      MobileNetV2-1.0 &  72.0\% &   5.78 &   2.89 &  1.34 &  0.67 &  0.34 &  0.17 &  0.17 \\
      MobileNetV2-1.4 &  74.7\% &   5.46 &   2.69 &  1.49 &  0.69 &  0.35 &  0.25 &  0.24 \\
          MnasNet-1.0 &  74.2\% &   5.84 &   2.92 &  1.34 &  0.72 &  0.34 &  0.19 &  0.17 \\
             DNANet-a &  77.1\% &  12.91 &   6.39 &  3.15 &  1.73 &  0.81 &  0.44 &  0.29 \\
             DNANet-b &  77.5\% &  13.31 &   6.14 &  3.02 &  1.66 &  0.78 &  0.40 &  0.37 \\
             DNANet-c &  77.8\% &  12.61 &   6.48 &  2.91 &  1.48 &  0.81 &  0.39 &  0.37 \\
             DNANet-d &  78.4\% &  13.04 &   7.08 &  3.43 &  1.75 &  0.83 &  0.58 &  0.54 \\
              DFNet-1 &  69.8\% &   3.59 &   1.72 &  0.90 &  0.43 &  0.24 &  0.11 &  0.07 \\
              DFNet-2 &  73.9\% &   6.69 &   3.41 &  1.54 &  0.86 &  0.40 &  0.22 &  0.12 \\
             DFNet-2a &  76.0\% &   7.93 &   4.40 &  2.01 &  1.00 &  0.50 &  0.26 &  0.19 \\
          OFANet-595M &  80.0\% &  12.30 &   6.20 &  3.08 &  1.69 &  0.83 &  0.45 &  0.41 \\
          OFANet-482M &  79.6\% &  13.28 &   6.19 &  3.07 &  1.66 &  0.83 &  0.40 &  0.33 \\
          OFANet-389M &  79.1\% &  11.83 &   5.89 &  2.73 &  1.49 &  0.74 &  0.38 &  0.26 \\
          OFANet-11ms &  76.1\% &   6.18 &   2.84 &  1.55 &  0.74 &  0.36 &  0.18 &  0.17 \\
           OFANet-9ms &  75.3\% &   5.04 &   2.68 &  1.35 &  0.62 &  0.32 &  0.18 &  0.14 \\
          GENet-light &  75.7\% &   5.48 &   2.94 &  1.37 &  0.69 &  0.34 &  0.19 &  0.09 \\
         GENet-normal &  80.0\% &   5.77 &   3.14 &  1.48 &  0.78 &  0.37 &  0.23 &  0.20 \\
          GENet-large &  81.3\% &   7.03 &   3.55 &  1.78 &  0.89 &  0.45 &  0.38 &  0.33 \\
    \bottomrule
    \end{tabular}
\end{table}

\begin{table}[hb]
  \centering
  \caption{Network latency (ms/image) on T4 TensorRT, FP16, batch size 1 to 32.}
  \label{tab:Network-latency-on-T4-TensorRT-FP16-all-batchsize}
  \begin{tabular}{llllllll}
    \toprule
                model &    acc &     1 &     2 &     4 &     8 &    16 &    32 \\
    \midrule
        RegNetY-200MF &  70.4\% &  4.95 &  2.60 &  1.46 &  0.65 &  0.39 &  0.28 \\
        RegNetY-400MF &  74.1\% &  4.22 &  1.96 &  1.04 &  0.64 &  0.43 &  0.34 \\
        RegNetY-600MF &  75.5\% &  2.80 &  1.42 &  0.78 &  0.48 &  0.34 &  0.28 \\
        RegNetY-800MF &  76.3\% &  2.54 &  1.39 &  0.76 &  0.48 &  0.35 &  0.30 \\
            ResNet-18 &  70.9\% &  1.43 &  0.83 &  0.56 &  0.35 &  0.26 &  0.21 \\
            ResNet-34 &  74.4\% &  2.16 &  1.17 &  0.80 &  0.49 &  0.38 &  0.33 \\
            ResNet-50 &  77.4\% &  2.27 &  1.31 &  0.86 &  0.61 &  0.51 &  0.46 \\
           ResNet-101 &  78.3\% &  3.61 &  2.13 &  1.36 &  0.96 &  0.82 &  0.74 \\
           ResNet-152 &  79.2\% &  4.92 &  2.88 &  1.85 &  1.32 &  1.15 &  1.04 \\
      EfficientNet-B0 &  76.3\% &  2.99 &  1.72 &  1.11 &  0.86 &  0.74 &  0.69 \\
      EfficientNet-B1 &  78.8\% &  4.44 &  2.54 &  1.66 &  1.30 &  1.15 &  1.08 \\
      EfficientNet-B2 &  79.8\% &  4.76 &  2.75 &  1.86 &  1.53 &  1.39 &  1.32 \\
      EfficientNet-B3 &  81.1\% &  6.03 &  3.74 &  2.85 &  2.60 &  2.44 &  2.36 \\
     MobileNetV2-0.25 &  51.8\% &  1.25 &  0.61 &  0.35 &  0.23 &  0.16 &  0.13 \\
      MobileNetV2-0.5 &  64.4\% &  1.21 &  0.67 &  0.39 &  0.25 &  0.19 &  0.16 \\
     MobileNetV2-0.75 &  69.4\% &  1.32 &  0.73 &  0.45 &  0.30 &  0.24 &  0.21 \\
      MobileNetV2-1.0 &  72.0\% &  1.36 &  0.78 &  0.48 &  0.32 &  0.25 &  0.22 \\
      MobileNetV2-1.4 &  74.7\% &  1.60 &  0.94 &  0.57 &  0.40 &  0.33 &  0.30 \\
          MnasNet-1.0 &  74.2\% &  1.49 &  0.84 &  0.52 &  0.34 &  0.28 &  0.24 \\
             DNANet-a &  77.1\% &  3.94 &  2.15 &  1.27 &  0.87 &  0.70 &  0.62 \\
             DNANet-b &  77.5\% &  3.74 &  2.12 &  1.30 &  0.90 &  0.73 &  0.66 \\
             DNANet-c &  77.8\% &  3.76 &  2.17 &  1.32 &  0.94 &  0.80 &  0.72 \\
             DNANet-d &  78.4\% &  4.20 &  2.52 &  1.60 &  1.17 &  1.01 &  0.95 \\
              DFNet-1 &  69.8\% &  1.42 &  0.74 &  0.46 &  0.29 &  0.18 &  0.14 \\
              DFNet-2 &  73.9\% &  2.05 &  1.12 &  0.66 &  0.41 &  0.26 &  0.21 \\
             DFNet-2a &  76.0\% &  2.18 &  1.25 &  0.70 &  0.41 &  0.28 &  0.24 \\
          OFANet-595M &  80.0\% &  3.06 &  1.71 &  1.13 &  0.88 &  0.77 &  0.71 \\
          OFANet-482M &  79.6\% &  2.93 &  1.65 &  1.02 &  0.75 &  0.63 &  0.58 \\
          OFANet-389M &  79.1\% &  2.62 &  1.45 &  0.89 &  0.64 &  0.53 &  0.48 \\
          OFANet-11ms &  76.1\% &  1.57 &  0.91 &  0.54 &  0.36 &  0.28 &  0.24 \\
           OFANet-9ms &  75.3\% &  1.39 &  0.76 &  0.48 &  0.31 &  0.24 &  0.21 \\
          GENet-light &  75.7\% &  1.66 &  0.90 &  0.48 &  0.28 &  0.18 &  0.14 \\
         GENet-normal &  80.0\% &  1.91 &  1.03 &  0.58 &  0.37 &  0.28 &  0.25 \\
          GENet-large &  81.3\% &  2.44 &  1.34 &  0.81 &  0.58 &  0.48 &  0.44 \\
    \bottomrule
    \end{tabular}
\end{table}

\begin{table}[hb]
  \centering
  \caption{Network latency (ms/image) on T4 TensorRT, INT8, batch size 1 to 32.}
  \label{tab:Network-latency-on-T4-TensorRT-INT8-all-batchsize}
  \begin{tabular}{llllllll}
    \toprule
                model &    acc &     1 &     2 &     4 &     8 &    16 &    32 \\
    \midrule
        RegNetY-200MF &  70.4\% &  2.08 &  1.07 &  0.56 &  0.34 &  0.21 &  0.14 \\
        RegNetY-400MF &  74.1\% &  2.35 &  1.27 &  0.73 &  0.43 &  0.28 &  0.20 \\
        RegNetY-600MF &  75.5\% &  2.41 &  1.26 &  0.74 &  0.43 &  0.29 &  0.22 \\
        RegNetY-800MF &  76.3\% &  2.26 &  1.27 &  0.73 &  0.44 &  0.30 &  0.23 \\
            ResNet-18 &  70.9\% &  1.24 &  0.68 &  0.41 &  0.24 &  0.16 &  0.13 \\
            ResNet-34 &  74.4\% &  1.78 &  1.04 &  0.56 &  0.34 &  0.23 &  0.19 \\
            ResNet-50 &  77.4\% &  1.83 &  1.04 &  0.60 &  0.40 &  0.30 &  0.26 \\
           ResNet-101 &  78.3\% &  2.67 &  1.59 &  0.95 &  0.61 &  0.47 &  0.41 \\
           ResNet-152 &  79.2\% &  3.49 &  2.21 &  1.24 &  0.81 &  0.65 &  0.55 \\
      EfficientNet-B0 &  76.3\% &  3.47 &  2.06 &  1.23 &  0.85 &  0.64 &  0.55 \\
      EfficientNet-B1 &  78.8\% &  5.39 &  3.17 &  1.93 &  1.32 &  0.99 &  0.85 \\
      EfficientNet-B2 &  79.8\% &  5.59 &  3.36 &  2.07 &  1.43 &  1.18 &  1.02 \\
      EfficientNet-B3 &  81.1\% &  7.03 &  4.45 &  2.94 &  2.22 &  1.88 &  1.67 \\
     MobileNetV2-0.25 &  51.8\% &  1.12 &  0.64 &  0.34 &  0.21 &  0.14 &  0.11 \\
      MobileNetV2-0.5 &  64.4\% &  1.04 &  0.65 &  0.35 &  0.22 &  0.16 &  0.13 \\
     MobileNetV2-0.75 &  69.4\% &  1.19 &  0.66 &  0.41 &  0.26 &  0.18 &  0.15 \\
      MobileNetV2-1.0 &  72.0\% &  1.16 &  0.77 &  0.42 &  0.26 &  0.19 &  0.16 \\
      MobileNetV2-1.4 &  74.7\% &  1.39 &  0.80 &  0.48 &  0.32 &  0.24 &  0.20 \\
          MnasNet-1.0 &  74.2\% &  1.29 &  0.76 &  0.45 &  0.28 &  0.21 &  0.18 \\
             DNANet-a &  77.1\% &  4.76 &  2.68 &  1.52 &  0.95 &  0.69 &  0.55 \\
             DNANet-b &  77.5\% &  4.65 &  2.73 &  1.61 &  1.06 &  0.80 &  0.66 \\
             DNANet-c &  77.8\% &  4.53 &  2.72 &  1.60 &  1.02 &  0.80 &  0.67 \\
             DNANet-d &  78.4\% &  5.39 &  3.21 &  1.99 &  1.39 &  1.11 &  0.97 \\
              DFNet-1 &  69.8\% &  1.24 &  0.71 &  0.42 &  0.23 &  0.15 &  0.11 \\
              DFNet-2 &  73.9\% &  1.85 &  1.04 &  0.58 &  0.31 &  0.19 &  0.15 \\
             DFNet-2a &  76.0\% &  1.80 &  1.07 &  0.58 &  0.33 &  0.21 &  0.16 \\
          OFANet-595M &  80.0\% &  4.28 &  2.40 &  1.39 &  0.93 &  0.71 &  0.63 \\
          OFANet-482M &  79.6\% &  4.25 &  2.31 &  1.33 &  0.85 &  0.62 &  0.52 \\
          OFANet-389M &  79.1\% &  3.86 &  2.00 &  1.16 &  0.73 &  0.52 &  0.43 \\
          OFANet-11ms &  76.1\% &  1.51 &  0.81 &  0.46 &  0.29 &  0.21 &  0.18 \\
           OFANet-9ms &  75.3\% &  1.23 &  0.73 &  0.41 &  0.26 &  0.18 &  0.15 \\
          GENet-light &  75.7\% &  1.42 &  0.79 &  0.41 &  0.24 &  0.15 &  0.10 \\
         GENet-normal &  80.0\% &  1.56 &  0.84 &  0.47 &  0.28 &  0.19 &  0.15 \\
          GENet-large &  81.3\% &  1.94 &  1.06 &  0.62 &  0.38 &  0.30 &  0.26 \\
    \bottomrule
    \end{tabular}
    
\end{table}

\end{document}